\pdfoutput=1

\documentclass[11pt]{article}

\usepackage{acl}

\usepackage{times}
\usepackage{latexsym}

\usepackage[T1]{fontenc}

\usepackage[utf8]{inputenc}

\usepackage{microtype}

\usepackage[export]{adjustbox}
\usepackage{amsmath}
\usepackage{amssymb}
\usepackage{booktabs}
\usepackage{bbm}
\usepackage{CJKutf8} 
\usepackage{enumitem}
\usepackage{float}
\usepackage{gb4e}\noautomath
\usepackage{graphicx}
\usepackage{geometry}
\usepackage{inconsolata}
\usepackage{multicol}
\usepackage{multirow}
\usepackage[smallerops,varbb]{newtxmath}
\usepackage{nicefrac}
\usepackage[colorinlistoftodos,prependcaption,textsize=tiny]{todonotes}
\usepackage{xspace}
\usepackage{arabtex}
\usepackage{tipa}

\renewenvironment{abstract}%
         {\centerline{\large\bf Abstract}%
          \begin{list}{}%
             {\setlength{\rightmargin}{0.6cm}%
              \setlength{\leftmargin}{0.6cm}}%
           \item[]\ignorespaces}%
         {\unskip\end{list}}

\newcommand{\mm}{\textsc{MultiMUC}\xspace}
\newcommand{\muc}{\textsc{MUC-4}\xspace}

\newcommand{\arson}{\texttt{arson}\xspace}
\newcommand{\attack}{\texttt{attack}\xspace}
\newcommand{\bombing}{\texttt{bombing}\xspace}
\newcommand{\kidnapping}{\texttt{kidnapping}\xspace}
\newcommand{\fws}{\texttt{forced work stoppage}\xspace}
\newcommand{\robbery}{\texttt{robbery}\xspace}

\newcommand{\gtt}{\textsc{GTT}\xspace}
\newcommand{\iterx}{\textsc{IterX}\xspace}
\newcommand{\multimuc}{\textsc{MultiMUC}\xspace}
\newcommand{\chatgpt}{\textsc{ChatGPT}\xspace}

\newcommand{\corresponding}{\textsuperscript{$\ast$}}

\newcommand{\pcomment}[1]{\textcolor{violet}{\emph{// #1}}}

\newcommand{\pref}[1]{(\ref{#1})}
\newcounter{mysub}
\makeatletter

\newcommand{\exs}{\save@counters\refstepcounter{mysub}\renewcommand{\thexnumi}{\arabic{xnumi}\alph{mysub}}\@ifnextchar [{\@ex}{\item}\reset@counters}
\makeatother

\newcommand{\ur}{\textsuperscript{\rm 1}}
\newcommand{\gtown}{\textsuperscript{\rm 2}}
\newcommand{\jhu}{\textsuperscript{\rm 3}}

\newcommand{\tgtauto}{$\textsc{Tgt}_\textsc{auto}$\xspace}
\newcommand{\tgtman}{$\textsc{Tgt}_\textsc{man}$\xspace}
\newcommand{\biman}{$\textsc{Bi}_\textsc{man}$\xspace}
\newcommand{\allman}{$\textsc{All}_\textsc{man}$\xspace}

\title{\mm: Multilingual Template Filling on MUC-4}

\renewcommand{\thefootnote}{$\text{*}$} 

\author{William Gantt\ur\corresponding \quad Shabnam Behzad\gtown \quad Hannah YoungEun An\ur \quad Yunmo Chen\jhu \\
        {\bf Aaron Steven White}\ur \quad {\bf Benjamin Van Durme}\jhu \quad {\bf Mahsa Yarmohammadi}\jhu\corresponding \\
        \ur~University of Rochester \quad \gtown~Georgetown University \quad \jhu~Johns Hopkins University \\
        \texttt{wgantt@cs.rochester.edu \quad mahsa@jhu.edu}}

\begin{document}
\maketitle

\begin{abstract}
\footnotetext{Corresponding authors}
\renewcommand{\thefootnote}{\arabic{footnote}}

We introduce \mm, the first multilingual parallel corpus for template filling, comprising translations of the classic \muc template filling benchmark into five languages: Arabic, Chinese, Farsi, Korean, and Russian. We obtain automatic translations from a strong multilingual machine translation system and manually project the original English annotations into each target language. For all languages, we also provide human translations for sentences in the dev and test splits that contain annotated template arguments. Finally, we present baselines on \mm both with state-of-the-art template filling models and with ChatGPT.
\end{abstract}
\renewcommand{\thefootnote}{\arabic{footnote}}

\section{Introduction}
\label{sec:intro}
The Message Understanding Conferences (MUCs) were a series of U.S. government-sponsored competitions that ran from the late 1980s through the late 1990s whose aim was to promote the development of systems for extracting complex relations from text, and which have been credited with inaugurating the field of information extraction \citep[IE;][]{grishman-sundheim-1996-message, grishman2019twenty}. The third MUC (MUC-3) introduced the now classic task of \emph{template filling}, in which systems must identify events, represented by predefined schemas or \emph{templates}, in a document, and populate roles or \emph{slots} in those templates with relevant information extracted or inferred from the text \citep{muc-1991-message}. MUC-3 focused on identifying various forms of terrorism (e.g.\ bombings, kidnappings) in news reports from numerous countries in Latin America. Systems were required to extract one template per incident, containing details about the perpetrators, their victims, the weapons used, and damaged physical infrastructure. The data, task specification, and evaluation methodology of MUC-3 were then refined and updated in MUC-4 \citep{muc-1992-message}.

\begin{figure}[t]
    \centering
    \includegraphics[width=\linewidth]{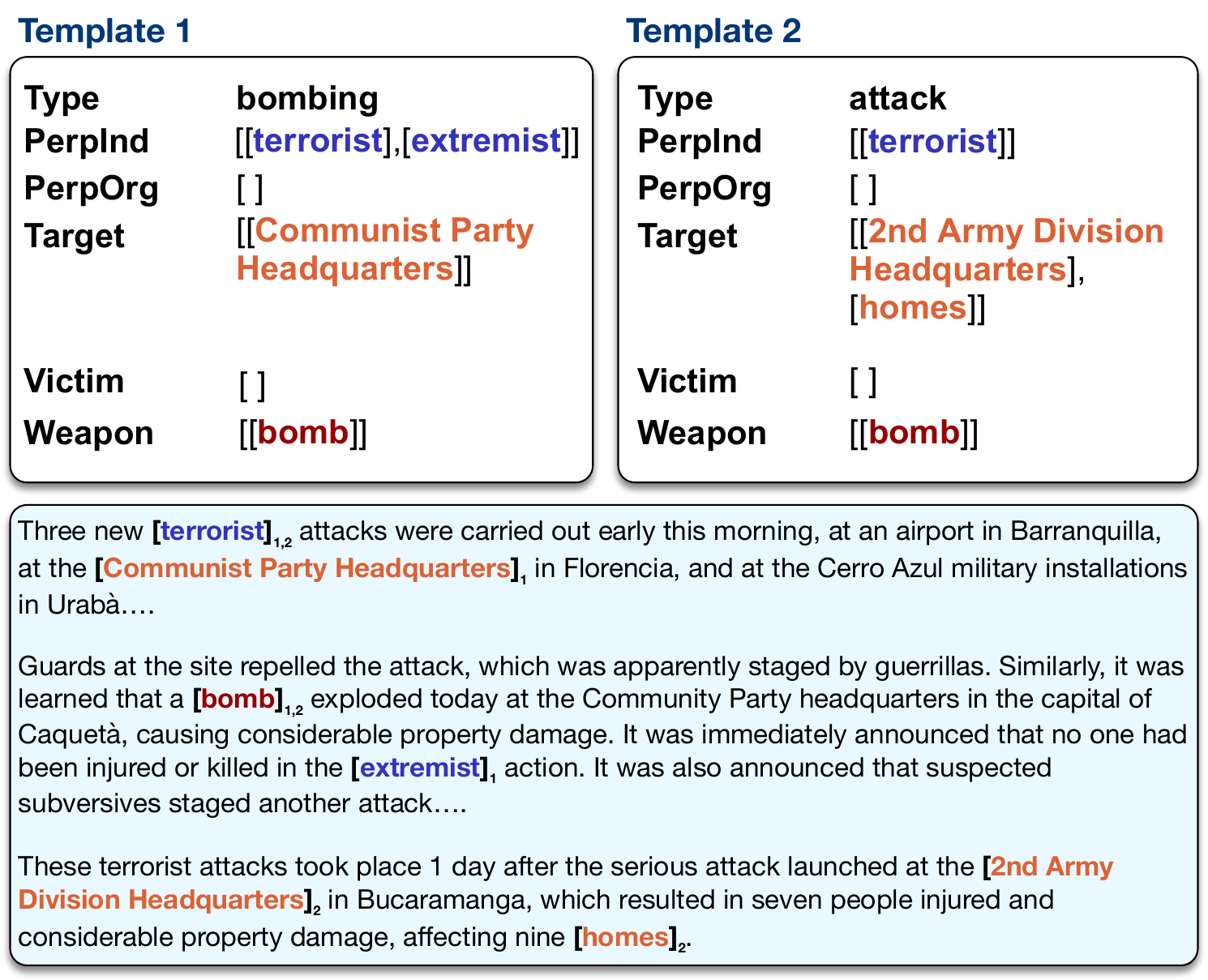}
    \caption{An excerpted document and its (simplified) gold templates from the \muc dataset.\vspace{-5mm}}
    \label{fig:Eng_example_doc}
\end{figure}

Since then, the MUC-4 corpus has been an enduring and productive driver of IE research---not only for template filling \citep{du-etal-2021-template, das-etal-2022-automatic, chen-etal-2023-iterative} and role-filler entity extraction \citep{patwardhan-riloff-2007-effective, patwardhan-riloff-2009-unified, huang-etal-2021-document, du-etal-2021-grit}, but also for template \emph{induction} \citep{chambers-jurafsky-2011-template, cheung-etal-2013-probabilistic}. Despite its international focus, MUC-4 is English-only, and multilingual, document-level IE datasets remain scarce. This work bolsters those resources with \textbf{\mm}, the first ever translations of the MUC-4 dataset, and to our knowledge the first multilingual \emph{parallel} corpus for template filling. This work provides:
\begin{itemize}
    \item High-quality, automatic translations of the MUC-4 dataset into five languages: Arabic, Chinese, Farsi, Korean, and Russian, along with (1) manual projections of the template annotations into each target language, and (2) human translations for sentences in the dev and test splits containing annotated arguments.
    \item Strong monolingual and bilingual supervised baselines for all five languages, based on state-of-the-art template filling models.
    \item Baselines for few-shot template filling with ChatGPT\footnote{\url{https://openai.com/blog/chatgpt}}---to our knowledge, the first few-shot evaluations of this task in the literature.
    \item Discussion and analysis of the translations, annotations, and model errors.
\end{itemize}

\noindent We release \mm and to help further research in multilingual, document-level IE.\footnote{\url{https://github.com/wgantt/multimuc}}

\section{Task and Corpus}
\label{sec:task}
\paragraph{Task} Formally, the template filling task takes the following inputs:
\begin{itemize}
    \itemsep 0.5pt
    \item A document $D$
    \item A template ontology $(\mathcal{T}, \mathcal{S})$, consisting of a set of template types $\mathcal{T} = \{T_1, ..., T_M\}$, each representing a distinct event type, as well as a set of $N_t$ slots for each template type $t \in \mathcal{T}$, representing the roles for that event type: $\mathcal{S} = \{S_t = \{s^{(1)}_t, \ldots, s^{(N_t)}_t\}: t \in \mathcal{T} \}$
\end{itemize}
Given $D$, systems must then determine the number of events or \emph{template instances} ($N_D \geq 0$) attested in $D$ (\textbf{template identification}), and populate the slots in each instance based on the information contained in $D$ about the event it represents (\textbf{slot filling}).\footnote{Following prior work \citep[\emph{i.a.}]{du-etal-2021-template, chen-etal-2023-iterative}, we will refer to template instances simply as \emph{templates}.} Note that $N_D$ is not given as input and may be zero; thus, part of the task is determining the \emph{relevancy} of a document given the ontology. Supposing template instance $i_j \in \{i_1,\ldots,i_{N_D}\}$ has type $t \in \mathcal{T}$, we can write $i_j = \{s^{(1)}_t: x^{(1)}, \ldots , s^{(N_t)}_t: x^{(N_t)}\}$, where $x^{(k)}$ is a (possibly null) filler of the appropriate type for slot $s^{(k)}_t$. In general, fillers may be of any type, though for \muc, they are constrained to two types in principle and just one in practice (see below).

\paragraph{Corpus} The MUC-4 corpus consists of 1,700 documents that concern incidents of terrorism and political violence in Latin America and that are annotated against a template ontology with six template types: \arson, \attack, \bombing, \kidnapping, \robbery, and \fws. Each template type is associated with the same set of 24 slots, which can be divided into \textbf{string-fill} slots---those that take (a set of) entities as fillers---and \textbf{set-fill} slots, which take a single filler from a fixed set of categorical values specific to each slot.\footnote{This is a minor simplification. See \autoref{app:muc_slots}.} \autoref{tab:dataset_stats} shows dataset statistics and \autoref{app:muc_slots} lists all slots.

\begin{table}[]
    \centering
    \begin{tabular}{l|ccc}
    \toprule
         & Train & Dev & Test \\
    \midrule
        Documents & 1300 & 200 & 200 \\
        Sentences & 18,317 & 2,989 & 2,702 \\
        Templates & 1,114 & 191 & 209 \\
    \bottomrule
    \end{tabular}
    \caption{Statistics for the \muc dataset. Sentence counts are based on our own sentence splits, as canonical sentence boundaries do not exist. Statistics are the same for all languages in \mm.\vspace{-5mm}}
    \label{tab:dataset_stats}
\end{table}

Since the original MUC evaluations, it has become standard to evaluate systems on simplified templates that contain only string-fill slots \citep[][\emph{i.a.}]{chambers-jurafsky-2011-template, du-etal-2021-grit, du-etal-2021-template, chen-etal-2023-iterative}, with the notable exception of the set-fill slot for template type. Additionally, while the gold data often lists multiple valid mentions for each entity filler, a system receives full credit for extracting just one of them. We follow both of these conventions in our work. The string-fill slots are \texttt{PerpInd} (individual perpetrators), \texttt{PerpOrg} (organizational perpetrators), \texttt{Target} (targeted infrastructure), \texttt{Weapon} (perpetrators' weapons), and \texttt{Victim} (victims of the event). \autoref{fig:Eng_example_doc} shows a \muc document and its simplified templates.

\section{Data Collection}
\label{sec:data_collection}
\begin{figure*}
\begin{center}
    \makebox[\textwidth]{\includegraphics[width=\textwidth]{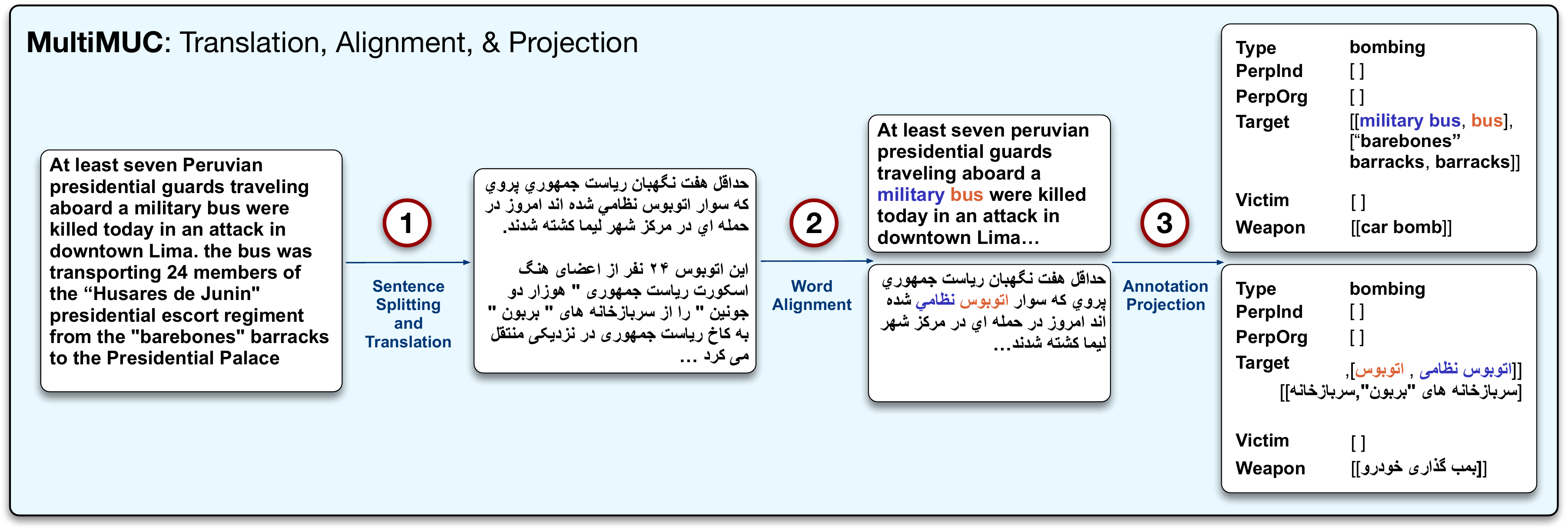}}%
    \caption{Process for creating projected target language data for \mm from the gold \muc data (\S\ref{sec:data_collection}). \vspace{-5mm}}
    \label{fig:data_projection}
\end{center}
\end{figure*}

The selection of languages for \mm was inspired by the five focal languages of the IARPA BETTER program\footnote{\url{https://www.iarpa.gov/index.php/research-programs/better}}, which introduced the Granular template filling task---a spiritual successor to MUC-4  \citep[see \S\ref{sec:background};][]{soboroff2023better}. For each language, our data collection process comprised four steps:
\begin{enumerate}
    \item \textbf{Preprocessing} of the \muc documents, including identification of sentence boundaries and locations of slot-filling entity mentions.
    \item \textbf{Machine Translation} of the documents into each of the five target languages.
    \item \textbf{Automatic Alignment} of slot-filling entity mentions in English with corresponding mentions in the target languages, followed by \emph{projection} of the template annotations.
    \item \textbf{Manual Correction} of entity mention alignments for all data splits, as well as translation corrections for sentences in the dev and test splits containing entity mentions.
\end{enumerate}
Each step is detailed separately below. \autoref{fig:data_projection} illustrates steps (1)-(3) for Farsi.

\subsection{Preprocessing}
We use the preprocessed version of the \muc dataset released by \citet{du-etal-2021-template}.\footnote{\url{https://github.com/xinyadu/gtt/}} Three quirks of the dataset deserve mention.

First, to our knowledge, the documents were never released with canonical sentence splits. As such, we used the NLTK Punkt sentence tokenizer \citep{bird2009natural}, to obtain sentence splits.\footnote{\url{https://www.nltk.org/_modules/nltk/tokenize/punkt.html}. Punkt is based on the unsupervised, multilingual sentence tokenization algorithm of \citet{kiss-strunk-2006-unsupervised}.}

Second, the text is uncased. This caused the sentence tokenizer to erroneously split a small number of sentences containing initialisms and titles (e.g.\ ``u.s.'' or ``dr.'') into two or more fragments. We manually corrected these cases by searching on a fixed set of problematic terms (identified via manual inspection) and combining identified fragments.\footnote{The terms were \emph{dr.}, \emph{mr.}, \emph{ms.}, \emph{mrs.}, \emph{gen.}, and \emph{u.s.}}

Third, character offsets of entity mentions are not annotated. This may be because evaluation has historically used string-based, rather than offset-based, matching to score string-fill slots. We follow \citet{du-etal-2021-template} in annotating the \emph{first} occurrence of each mention string in a document and leave annotation of later occurrences for future work.

\subsection{Machine Translation}\label{subsec:mt}
Given the preprocessed English text, we obtain automatic translations of all 1,700 MUC-4 documents for all five of the target languages. Our MT system has a Stratified Mixture of Experts (SMoE) architecture \cite{xu2023parameterefficient} for multilingual translation. Mixture-of-experts (MoE) \cite{conf/iclr/ShazeerMMDLHD17,DBLP:conf/iclr/LepikhinLXCFHKS21} significantly scales up the number of parameters of multilingual transformer-based MT models while maintaining low computational requirements per token. SMoE enhances MoE models by assigning dynamic model capacity to different tokens, thus enabling more efficient parameter use. SMoE has demonstrated improvements over state-of-the-art MoE baselines \cite{xu2023parameterefficient}.

We use an SMoE model pretrained on the primary bitexts of six languages from NLLB \cite{costa2022no}, covering over 70 million parallel sentences and all \mm languages.

\subsection{Automatic Alignment and Projection}\label{subsec:align}

Data projection involves automatically transferring span-level annotations from a source language to a target language based on word-to-word alignments. Given the translated documents, we first align each word in an English (source) sentence to the corresponding word(s) in the target sentence. Mentions in the target language are thus given by the sequence of target language tokens aligned to each token in an annotated source mention, and the corresponding slot and template in the source are thereby implicitly projected to the target.


We use Awesome-align \cite{dou-neubig-2021-word}, an embedding-based word aligner that derives word alignments via comparison of word embeddings. Awesome-align fine-tunes a pretrained language model \citep[in our case, XLM-R;][]{conneau-etal-2020-unsupervised} on parallel text or gold word alignments with training objectives designed to improve alignments.


We reuse the models and empirically chosen hyperparameters from prior work for a similar task \cite{multicoref}. These models are XLM-R encoders fine-tuned on around two million parallel target language-English sentences from the OSCAR corpus \cite{abadji-etal-2022-towards}. The encoders are further fine-tuned on gold alignments from  GALE Chinese–English \cite{Li2015GALECP}, and the Farsi-English corpus by \citet{Tavakoli-2014}, containing 2,800 Chinese–English and 1,200 Farsi-English sentence pairs with gold alignments. We further fine-tuned the Arabic model on the 2,300 GALE Arabic-English \cite{Li2013GALEAR} sentence pairs with gold alignments.

\subsection{Translation and Alignment Correction}\label{subsec:annotation}
While we find our automatic alignments to be of good quality (\autoref{tab:mm_stats}), prior work has shown that for some IE tasks, models can benefit meaningfully from access to gold alignments \citep{stengel-eskin-etal-2019-discriminative,behzad-etal-2023-effect}. Accordingly, we recruited annotators to inspect and (if necessary) correct the automatic alignments for all sentences containing the first occurrence of some entity mention. Additionally, for the dev and test splits, annotators corrected the \emph{translations} of these sentences.

Annotation was performed using the TASA annotation tool.\footnote{\url{https://github.com/hltcoe/tasa}} Annotators included students from Johns Hopkins and professional linguists from the Human Language Technology Center of Excellence (HLTCOE). All are either native speakers of the language they annotated or have extensive training in that language. For practice, annotators completed 10 tasks that were not included in the final data. Given the annotators' level of competence as well as budgetary constraints, only a single annotator completed each main task. Between one and four annotators worked on each language, with tasks distributed based on availability. Three of the annotators are authors of this work and were not paid; others were paid at an average rate of \$0.29 per task.\footnote{This figure is based on average pay for the student annotators. Linguists were paid by the HLTCOE at a rate that was not disclosed to the authors.} Task instructions, screenshots of the interface, and agreement statistics are in \autoref{app:data_collection}.

Entity and mention statistics for the training split of each language are shown in \autoref{tab:mm_stats}. In general, only a small fraction of the automatic alignments required correction: Even for the two languages requiring the most correction (Chinese and Russian) fully 77.4\% of target language mentions were unchanged from the automatic alignment. For the language requiring the least correction (Arabic), 86.5\% of spans were unchanged. This is testament to the quality of the alignments, though alignment quality is necessarily constrained by translation quality, which we discuss in \autoref{app:data_collection}.

\begin{table}[]
    \centering
    \adjustbox{max width=\linewidth}{
    \begin{tabular}{l|ccccc}
    \toprule
         & Ar & Fa & Ko & Ru & Zh \\
    \midrule
        Entities & 2,421 & 2,432 & 2,417 & 2,394 & 2,071 \\
        $\text{Mentions}_\text{man}$ & 3,074 & 3,136 & 3,076 & 3,019 & 2,597 \\
        $\enspace$ unchanged & 86.5 & 84.0 & 79.7 & 77.4 & 77.4 \\
    \bottomrule
    \end{tabular}
    }
    \caption{Entity and mention counts for the \mm training set. ``$\text{Mentions}_\text{man}$'' denotes \emph{annotated} mentions. ``Unchanged'' denotes the percentage of $\text{Mentions}_\text{man}$ unchanged from the automatic alignment.\vspace{-5mm}}
    \label{tab:mm_stats}
\end{table}

\section{Experiments}
\label{sec:experiments}
We present three sets of experiments. All make use of the following three variations on training and dev data, designed to assess both the impact of alignment corrections and of parallel data:

\begin{enumerate}
    \item \textbf{\tgtauto} uses only \emph{target} language data, with mentions obtained via \emph{automatic} alignments.
    \item \textbf{\tgtman} uses only \emph{target} language  data, but with the \emph{manually} corrected alignments for the training set and the corrected alignments \emph{and translations} for the dev set.
    \item \textbf{\biman} is the same as \tgtman, but adds gold English training data (yielding \emph{bilingual} data).
\end{enumerate}

In all experiments, we report results on the \emph{corrected} test set.

\subsection{Span Extraction}\label{subsec:span_extraction}
\paragraph{Setup} Prior work investigating the impact of alignment quality in IE has focused on span labeling tasks such as NER or SRL \citep{stengel-eskin-etal-2019-discriminative, behzad-etal-2023-effect}, as these tasks arguably give the most direct view on the downstream impact of improved alignments. In our first set of experiments, we follow this line of work and assess span extraction and labeling performance on \mm using the neural span extractor of \citet{xia-etal-2021-lome}, which has achieved state-of-the-art performance on FrameNet \citep{baker1998berkeley}. We train the system to extract all slot-filling entity mentions and to label them with their slot.

\paragraph{Results} Labeled and Unlabeled exact match $\text{F}_1$ scores for the three settings are shown in \autoref{tab:span_extraction}. Across almost all languages, we observe improvements on both metrics when training on corrected (\tgtman) vs.\ uncorrected (\tgtauto) data. Given that a fairly small proportion of spans in the data were changed between these settings, some of the gains may also be explained by access to corrected dev data in the \tgtman setting.

\begin{table}[]
    \centering
    \adjustbox{max width=\linewidth}{
    \begin{tabular}{l|ccccc}
    \toprule
         & Ar & Fa & Ko & Ru & Zh \\
    \midrule
        \textsc{\tgtauto} & 51.92 & 49.84 & 51.14 & 58.15 & \bf{54.46} \\
        \textsc{\tgtman} & \bf{56.25} & \bf{55.62} & 52.00 & \bf{59.34} & 52.88 \\
        \textsc{\biman} & 54.89 & 53.34 & \bf{55.41} & 57.40 & 53.44 \\
        \midrule
        \textsc{\tgtauto} & 54.62 & 52.07 & 52.86 & 60.05 & 55.51 \\
        \textsc{\tgtman} & \bf{58.88} & \bf{56.82} & 54.76 & \bf{62.54} & 54.64 \\
        \textsc{\biman} & 56.60 & 55.10 & \bf{57.78} & 59.66 & \bf{55.66} \\
    \bottomrule
    \end{tabular}
    }
    \caption{Labeled (top) and unlabeled (bottom) exact span match $\text{F}_1$ scores for all three data settings on the annotated test splits.\vspace{-5mm}}
    \label{tab:span_extraction}
\end{table}

\subsection{Template Filling with Fine-Tuned Models}\label{subsec:tf_models}
\paragraph{Setup} Our second set of experiments turns to template filling proper, focusing on the two models to have most recently achieved state-of-the-art on \muc. The first is \gtt \citep{du-etal-2021-template}, which uses a single BERT-base model \citep{devlin-etal-2019-bert} as both an encoder (to encode the document) and as a decoder, using causal masking and pointer decoding to generate linearized templates. As a minimal modification to support the \multimuc languages, we use \emph{m}BERT-base \citep{devlin-etal-2019-bert} in lieu of BERT-base, keeping all other aspects of the architecture unchanged.

The second model is \iterx \citep{chen-etal-2023-iterative}, which holds state-of-the-art on \muc. \iterx treats template filling as autoregressive span classification, assigning each of a set of candidate spans (extracted by an upstream system) either to a slot in the current template or else to a special ``null'' slot to indicate that the span fills \emph{no} slot in that template. Embeddings for the candidate spans are updated at each iteration based on their use in previous templates, and are used to condition the span assignments for subsequent templates. \citeauthor{chen-etal-2023-iterative} obtain their best \muc results with a T5 encoder \citep{raffel2020exploring}. As with \gtt, we make a minimal modification to the English base model by substituting \emph{m}T5-base \citep{xue-etal-2021-mt5} for the encoder, keeping all else unchanged.\footnote{We stress that our interest here is to present the best results for each model type and to evaluate cross-lingual performance variation \emph{within} type, not in cross-type comparisons. For a comparison on \muc of \iterx and \gtt under identical encoders, see \citet{chen-etal-2023-iterative}. Additional details on architectures and hyperparameters are provided in \autoref{app:training}.}

\paragraph{Evaluation} Evaluating template filling systems requires aligning predicted ($P$) and reference ($R$) templates, subject to the constraints that each reference template is aligned to at most one predicted one and that their types match. This is treated as a maximum bipartite matching problem, in which one seeks the alignment that yields a maximum total score over template pairs $(P,R)$ given some template similarity function $\phi_T$:
\begin{equation}
    A^* = \underset{A}{\text{argmax}} \sum_{(P,R) \in A} \phi_T(P,R)
\end{equation}\label{eq:optimal_alignment}

\noindent $\phi_T(P,R)$ measures similarity between two templates in terms of similarity of their slot fillers, and there are different ways to specify this. \citet{du-etal-2021-template} propose the CEAF-REE metric, which computes an optimal alignment between predicted and reference \emph{entities} similar to the CEAF metric for coreference resolution \citep{luo-2005-coreference}, but where aligned entities must fill the same slot. CEAF-REE selects the template alignment that yields the highest micro-$\text{F}_1$ over all slot fills, \emph{including template type}. However, \citet{chen-etal-2023-iterative} take issue with certain properties of CEAF-REE and propose a variant called CEAF-R\emph{M}E. The key differences from CEAF-REE are (1) the template type is \emph{excluded} from the $\text{F}_1$ calculation and (2) a different similarity function is used for entity alignments. We report both metrics and refer the reader to their paper or to \citet{chen-etal-2023-unified} for  details.\footnote{In \citeauthor{chen-etal-2023-iterative}'s terminology, we report $\text{CEAF-REE}_\text{impl}$ and $\text{CEAF-RME}_{\phi_3}$.}

\paragraph{Results} Results for all languages are presented in the first six rows of \autoref{tab:main_results}. Several observations stand out. First, for nearly all languages, both models obtain their strongest performance when trained jointly on English and target language data (\biman). This is consistent with past findings in IE establishing the value of English training data for lower-resource target languages \citep[][\emph{i.a.}]{subburathinam-etal-2019-cross, yarmohammadi-etal-2021-everything, fincke2022language}. While the impact of the English data is valuable for both models, it is especially so for \iterx, for which it boosts performance relative to the next best setting by an average of about 8.3 CEAF-REE $\text{F}_1$ and an average of over 4.7 CEAF-RME $\text{F}_1$ (compared to  3.2 and 2.6 $\text{F}_1$ for \gtt).\footnote{We additionally considered a fourth setting, \allman, in which models are jointly trained on the corrected data for all \mm languages, though this did not show clear gains over the \biman setting. See \autoref{app:additional_results}.}

Second, the benefits of training on the target language data with corrected alignments (\tgtman) are most evident for \gtt, for which it shows consistent improvements relative to no corrections (\tgtauto) for CEAF-RME scores.\footnote{CEAF-\emph{REE} scores are expected to show a noisier relationship with alignment correction due to the inclusion of the template type slot in the $\text{F}_1$ calculation, as accuracy is usually much higher for this slot than for others.} In contrast, performance does not substantially differ between the two settings for \iterx. This may be a consequence of \iterx's reliance on an upstream system for its candidate spans: to isolate the effect of \iterx \emph{training}, these candidates were fixed across settings at inference time, but it's plausible that the added value of corrected alignments lies chiefly in the span extraction step, prior to IterX training.

Lastly, the best scores for both models in all five \mm langauges are low compared to the best reported results on English. There is clear room for improvement across all languages, and we are excited by the prospect of future models better tailored to specific languages.


\begin{table*}[t]
    \centering
    \adjustbox{max width=\textwidth}{
    \begin{tabular}{ll|cccccc|cccccc}
    \toprule
    & & \multicolumn{6}{c}{\underline{CEAF-REE}} & \multicolumn{6}{c}{\underline{CEAF-RME}} \\
    & & En & Ar & Fa & Ko & Ru & Zh & En & Ar & Fa & Ko & Ru & Zh \\
        \midrule
    \multirow{4}{*}{\gtt} & \textsc{\tgtauto} & \multirow{4}{*}{50.23} & 24.26 & 31.46 & 34.17 & 35.38 & 36.74 & & 11.27 & 16.24 & 18.24 & 20.23 & 18.90 \\
    & \textsc{\tgtman} & & 28.81 & 36.01 & 33.79 & 38.05 & 36.35 & 32.30 & 15.05 & 21.27 & 18.71 & \textbf{22.44} & 19.11 \\
    & \textsc{\biman} & & 36.76 & \textbf{37.91} & 36.52 & 36.97 & \textbf{41.48} & & \textbf{21.98} & \textbf{22.44} & \textbf{20.71} & 21.26 & \textbf{23.26} \\
        \midrule
    \multirow{4}{*}{\iterx} & \textsc{\tgtauto} & \multirow{4}{*}{53.00} & 25.55 & 27.15 & 25.99 & 29.61 & 27.54 & \multirow{4}{*}{35.20} & 15.96 & 17.78 & 16.52 & 19.58 & 17.60 \\
    & \textsc{\tgtman} & & 25.70 & 25.36 & 27.24 & 30.08 & 27.32 & & 15.73 & 16.41 & 17.11 & 19.30 & 17.06 \\
    & \textsc{\biman} & & \textbf{34.73} & \textbf{33.15} & \textbf{37.02} & \textbf{36.95} & \textbf{36.02} & & \textbf{21.46} & \textbf{20.66} & \textbf{23.91} & \textbf{23.77} & \textbf{21.93} \\
        \midrule
    \multirow{2}{*}{\chatgpt} & \textsc{\tgtman} & \multirow{2}{*}{29.11} & 23.77 & 21.02 & \textbf{17.14} & \textbf{25.40} & 23.36 & \multirow{2}{*}{22.41} & 14.67 & 12.91 & \phantom{0}6.73 & \textbf{16.38} & \textbf{15.02} \\
    & \textsc{\biman} & & \textbf{24.62} & \textbf{22.06} & 16.85 & 24.90 & \textbf{24.46} & & \textbf{14.79} & \textbf{13.42} & \textbf{\phantom{0}7.12} & 15.36 & 13.99 \\
    \bottomrule
    \end{tabular}
    }
    \caption{CEAF-REE and CEAF-RME $\text{F}_1$ scores on English and the five \multimuc languages for \gtt \citep{du-etal-2021-template}, \iterx \citep{chen-etal-2023-iterative}, and \chatgpt under the data settings described in \S\ref{sec:experiments}. English results are the best ones reported by \citeauthor{chen-etal-2023-iterative}, except for \chatgpt, and do not correspond to any of the three data settings. \textbf{Bolded} results are best results within model type. See \S\ref{subsec:tf_models} for caveats about cross-type comparisons.\vspace{-5mm}}
    \label{tab:main_results}
\end{table*}

\subsection{Few-Shot Template Filling}\label{subsec:few_shot}
With the staggering leaps in the capabilities of large language models of the past couple years, an immediate question for most tasks asks how competitive these models are in a zero- or few-shot setting compared to smaller, fine-tuned models (\S\ref{subsec:tf_models}). We consider this question for \mm, investigating the capabilities of ChatGPT\footnote{\url{https://openai.com/blog/chatgpt}} on few-shot template filling. While ChatGPT's training corpus is predominantly English, some works have studied its abilities on MT \citep{jiao2023chatgpt, peng2023towards} and on IE tasks in other languages \citep{lai2023chatgpt}, and have found solid results. To our knowledge, this is the first work exploring few-shot template filling \emph{at all}.

\paragraph{Setup} We use the long-context version of ChatGPT (\texttt{gpt-3.5-turbo-16k-0613}) and evaluate in the \tgtman and \biman settings. The system prompt instructs the model to adopt the persona of an expert in IE and to perform extraction on a target document. The user prompt provides more detailed instructions, including the desired output format for extracted templates, as well as three examples of other documents with their gold templates.\footnote{Some effort was invested in identifying effective prompts for this task, but our aim here is merely a reasonable few-shot baseline---\emph{not} an extensive prompt engineering project. Prompt examples and hyperparameters are in \autoref{app:training}.} For the \tgtman setting, example documents are chosen from the target language training set using a BM25 retrieval model and are sorted so that the most relevant example is last. For the \biman setting, we replace the most relevant target language example with the same example in English.

\paragraph{Results} Results are shown in the bottom two rows of \autoref{tab:main_results}. Performance in both settings trails the performance of  \iterx and \gtt across languages---a finding in line with prior work showing that ChatGPT's few-shot capabilities on many tasks still fall short of those of the best supervised models \citep{lai2023chatgpt, gao2023exploring}, and an unsurprising result given its predominantly English training corpus. Furthermore, the clear gains from English training data for the supervised models do not clearly carry over here: Including a relevant English document in the prompt helps only in some cases and even then only modestly.

\section{Discussion}
\label{sec:discussion}
Here we present some analysis of model errors (\S\ref{subsec:model_errors}) and also discuss observations and challenges from annotation (\S\ref{subsec:annotation_challenges}).

\subsection{Model Errors}\label{subsec:model_errors}
We use the template filling error analysis tool of \citet{das-etal-2022-automatic} to understand the distribution of error types in the predictions from \gtt.\footnote{\url{https://github.com/IceJinx33/auto-err-template-fill/}} \citeauthor{das-etal-2022-automatic} define a set of transformations by which a set of predicted templates may be converted into the gold ones, given an optimized template alignment (see \S\ref{sec:experiments}). These include insertion and deletion transformations for templates and role fillers, as well as edit transformations for mentions and their role assignments. Error types are then defined in terms of \emph{sequences} of these transformations.

\begin{figure*}[t]
\begin{center}
    \makebox[\textwidth]{\includegraphics[width=\textwidth]{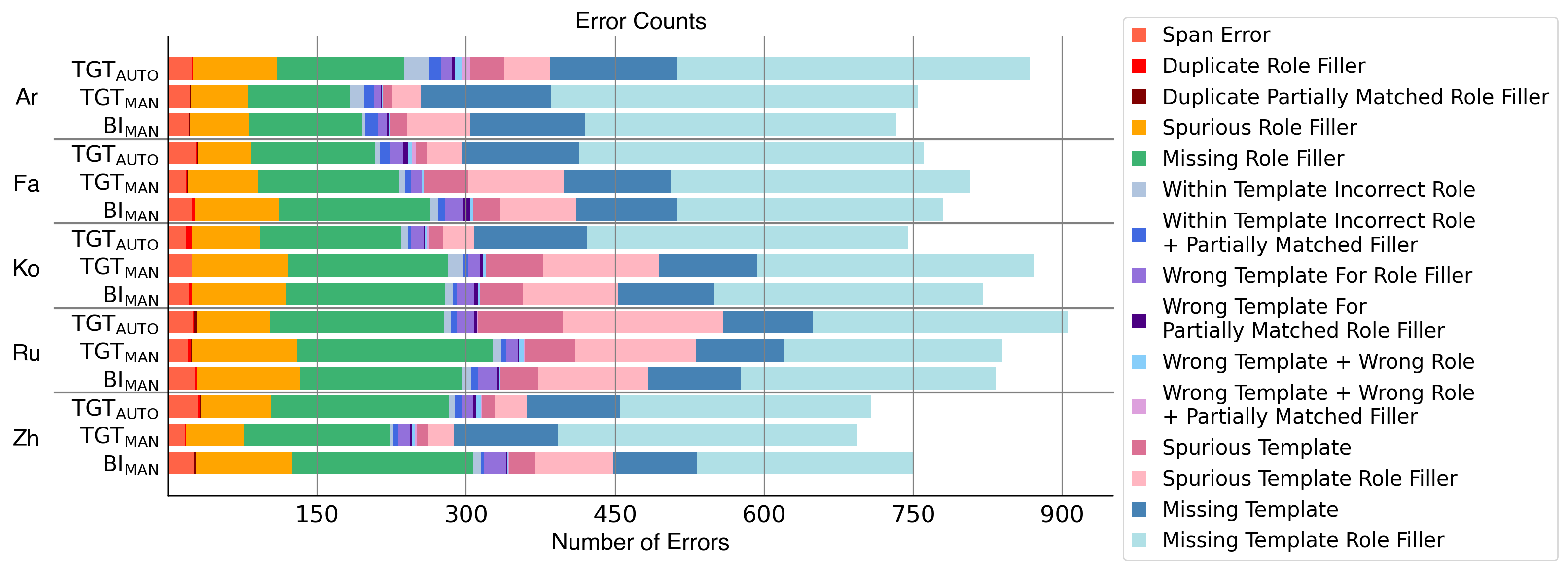}}%
    \caption{Automated error analysis results based on the tool provided by \citet{das-etal-2022-automatic} for \gtt test set predictions for all \mm languages and all data settings (see \S\ref{sec:experiments}). Missing role filler errors predominate.\vspace{-5mm}}
    \label{fig:error_count}
\end{center}

\end{figure*}

\autoref{fig:error_count} shows a breakdown of errors by type for all languages and all three data settings for \gtt. Consistent with \citeauthor{das-etal-2022-automatic}'s observations for \muc, we find that, across languages and settings, missing role fillers account for a majority of the errors.\footnote{This includes both ``Missing Role Filler'' errors (i.e.\ role fillers missing from a predicted template) and ``Missing Template Role Filler'' errors (i.e.\ role fillers missing due to the associated template not being predicted in the first place).} This is unsurprising when considering both that \gtt's extractions heavily favor precision \citep{du-etal-2021-template} and that models tend to struggle significantly with template recall, perhaps due to difficulty in \emph{individuating} events \citep{gantt-etal-2023-event}. Spurious templates and role fillers represent a smaller but non-trivial fraction of all errors.

\subsection{Annotation Observations}\label{subsec:annotation_challenges}
We now discuss observations and challenges from the annotation process. While there are obviously many language-specific considerations for both translation and alignment, we highlight several that were common to two or more languages.
\subsubsection{Proper Nouns}
\begin{CJK}{UTF8}{gbsn}
\muc annotations contain a significant number of proper nouns with a single canonical form, and these were sometimes translated into multiple forms in the target language, including both acceptable variants (e.g.\ the Farsi {\small``\setfarsi\novocalize \<hotel ^serAton>''} [\textipa{hoh-tel she-raa-tohn}] or {\small``\setfarsi\novocalize \<hotel ^serAyton>''} [\textipa{hoh-tel she-reye-tohn}] for \emph{Sheraton Hotel}) and orthographic errors ( \CJKfamily{mj}레이 [\textipa{\:le.i}], 릴리 [\textipa{\:li\:l.\:li}], or 릴 [\textipa{\:li\:l}] for the name \emph{Leigh}). In Chinese, each syllable in a proper noun may be translated into one of several characters that approximate the pronunciation. E.g.\ the first syllable of \emph{Guatemala} may phonetically correspond to 危 [wēi] or 瓜 [guā], and the noun as a whole can be translated as either 危地马拉 or 瓜地马拉. These forms were canonicalized as much as possible in the dev and test annotations, but this could not be done for the training set, for which only span alignments were corrected.
\end{CJK}

\subsubsection{Word Order}
In general, Farsi has subject-object-verb (SOV) word order and Arabic has verb-subject-object (VSO) order. However, in both languages, the order can sometimes change due to context, certain case endings, and adverbs. In a number of instances, annotators noted that the automatic translations use the standard word order even when changing it would result in a more natural phrasing. As an example, for the sentence ``the rebels who (...) attacked the building'', the automatic Arabic translation was {\small``\setfarsi\novocalize \<hAjem almtmardUn ala_dIn (...) almabnI>''}, where {\small``\setfarsi\novocalize \<hAjem>''} is the verb, {\small``\setfarsi\novocalize \<almtmardUn>''} is the subject and {\small``\setfarsi\novocalize \<almabnI>''} is the object. But a more natural translation would be {\small``\setfarsi\novocalize \<almtmardUn ala_dIn (...) hAjemUA almabnI>''}. Such cases were corrected in dev and test.

\subsubsection{Numeral classifiers}
\begin{CJK}{UTF8}{} \CJKfamily{mj} 
Chinese and Korean mark nouns with classifiers (CL) when naming and counting them. In both languages, a CL always follows a numeral when an explicit number is present, and in Korean, when the combination of a numeral and a CL follows its associated noun, aligning the classifier to the noun is less desirable, as this yields discontiguous target language spans. As such, annotators aligned numerals in English to both the numeral and CL in the target languages, as illustrated in Example \pref{korean-ex-cl}. Relatedly, for Chinese translation correction, annotators combined a (numeral, CL) pair into one token when they were translated as separate tokens.

\begin{exe}
    \ex \label{korean-ex-cl} 
    경찰 \qquad\enspace\thinspace 세 \quad\medspace\thinspace 명 \qquad\quad(Korean)\\
    \textit{gyeongchal} \thinspace \textit{\textbf{se}} \quad\thickspace\medspace \textbf{myeong}\\
    policeman \thickspace\medspace \textbf{three} \medspace \textbf{CL} \\
    `\textbf{three} policemen'
\end{exe}
\end{CJK}

\section{Related Work}
\label{sec:background}
\paragraph{Template Filling} Template filling has a long history. Participants in the MUCs, starting with MUC-3 \citep{muc-1991-message} and MUC-4 \citep{muc-1992-message}, largely developed pipelined, rule-based systems with individual modules designed to solve problems that are now major NLP tasks in their own right, such as coreference resolution and semantic role labeling \citep{hobbs-1993-generic, grishman2019twenty}. MUC-5 introduced a considerably more complicated template ontology that represented entities \emph{themselves} as templates, yielding nested template structures \citep{muc-1993-message}. MUC-6 \citep{muc-1995-message} and MUC-7 \citep{muc-1998-message} also had nested templates, but the entity templates were pared down to fewer slots and their ontologies had only a single event type.

Following the MUCs, many works revisiting these corpora focused on \emph{role-filler entity extraction}, a simplified form of template filling in which the goal is to identify all entity fillers, but without collating them into distinct templates \citep{patwardhan-riloff-2007-effective, patwardhan-riloff-2009-unified, huang-riloff-2011-peeling, huang2012modeling, du-etal-2021-grit, huang-etal-2021-document}. Template filling also differs from two other, closely related tasks. First, it differs from document-level $N$-\emph{ary relation extraction} in being \emph{event}-centric and in permitting null arguments. Second, it differs from \emph{event extraction} (EE) in not requiring extraction of event triggers (indeed, MUC-4 does not annotate triggers).


\paragraph{Multilingual Template Filling} Works cited in preceding sections \citep{du-etal-2021-template, chen-etal-2023-iterative, das-etal-2022-automatic} exhaust deep learning-era efforts on template filling with \muc. Even as early as the \muc conference itself, though, there was interest in extending template filling systems to other languages. NYU's PROTEUS system, for instance, was extended to handle Spanish documents \citep{grishman-etal-1992-new}, and the SOLOMON system from Systems Research and Applications (SRA) was enhanced to handle both Spanish and Japanese documents \citep{aone-etal-1992-sra, aone-etal-1993-murasaki}. This work presaged MUC-5, which had evaluations in both English and Japanese, but as best we know, no corpora were ever released for either language.

A number of multilingual resources exist for \emph{sentence-level} event extraction, such as ACE \citep[in Arabic, Chinese, and English;][]{doddington-etal-2004-automatic, walker-etal-2006-multilingual} and the Light and Rich ERE datasets from the DARPA DEFT program \citep[Chinese, English, and Spanish;][]{song-etal-2015-light}, though analogous resources at the document level are much more limited. The primary resource of note here is the Granular dataset from the IARPA BETTER program \citep{soboroff2023better}, featuring an ontology of six diverse template types (e.g.\ protests, epidemics, natural disasters), and covering news articles in English and five other languages. Granular is notable as the only multilingual template filling dataset that has both gold document texts and gold template annotations, though this is not parallel data and the corpus is much smaller than \muc, with only several hundred documents.



\paragraph{Cross-Lingual Alignment and Projection}

Cross-lingual projection is a method for transferring annotations from a source language to a target language, used primarily to create cross-lingual datasets for structured prediction tasks~\cite[\emph{i.a.}]{yarowsky-ngai-2001-inducing,aminian-etal-2019-cross,fei-etal-2020-cross,daza-frank-2020-x,ozaki-etal-2021-project,yarmohammadi-etal-2021-everything,chen-etal-2023-frustratingly}. The approach relies on two main steps: translation and source-to-target word alignment, and thus relies on high-quality translations and alignments between source and target texts. Studies have shown that access to gold entity alignments can improve downstream results \cite{stengel-eskin-etal-2019-discriminative,behzad-etal-2023-effect}.

\section{Conclusion}
\label{sec:conclusion}
We have introduced \mm---the first multilingual \emph{parallel} template filling dataset, featuring high-quality automatic translations of the \muc corpus along with human translations of key portions of the dev and test splits, and human-annotated alignments for all fillers of string-fill slots. Moreover, we have established strong mono- and bilingual baselines using two recent, top-performing template filling models, as well as baselines for few-shot template filling---to our knowledge, the first few-shot evaluations for this task. Lastly, we have highlighted some observations and challenges involved in constructing this resource and presented a detailed breakdown of model errors. We hope that this work will facilitate further research on multilingual IE at the document level.

\section*{Limitations}
\label{sec:limitations}
Ideally, all datasets that include machine-generated outputs would have exhaustive human verification and correction of those outputs. This of course applies to \mm: while the dataset provides human translations of key portions of the dev and test splits (all sentences containing the first occurrence of each entity mention), the majority of sentences in the dataset are machine-translated, which results in a small number of data projection failures (see \autoref{app:data_collection}). Obtaining gold translations and entity alignments for the entire corpus was simply infeasible with the personnel and budget available to us for the present work. Regardless, the automatic alignments and translations are of good quality (see \S\ref{sec:data_collection} and \autoref{app:data_collection}) and make \mm a valuable resource for developing document-level IE systems in multiple languages.


\section*{Ethics Statement}
\label{sec:ethics}
While the MUC-4 dataset has an established history in the NLP and IE communities, the documents it contains---and \mm, by extension---concern historical incidents of terrorism and use the names of real persons involved in those incidents. Caution is therefore warranted in using this data in the training, development, or deployment of models for template filling or for other tasks. Given the difficulty of template filling, even the best current systems trained to perform this task will hallucinate or misrepresent a non-trivial portion of the events they extract.

\section*{Acknowledgments}
\label{sec:acknowledgments}
\mm was made possible by the annotation work of Amir Hussein and Yefu Wang at Johns Hopkins, together with a dedicated team of linguists at the HLTCOE, which was led by Rachel Hare and which included (in alphabetical order by last name) Justin Boone, Alyssa Cabrera, Celia Colon-Fernandez, Everest Guerrero, Jacqueline Froelich, Jaymin Ko, Bethany Osterman, Faith Rodriguez, and Christopher Schaff. Our work was supported by a fellowship from JHU + Amazon Initiative for Interactive AI (AI2AI) and by the IARPA BETTER program (201919051600005). We also thank Craig Harman for assistance with the annotation setup and Hyokun Yun for helpful discussions.

\bibliography{anthology,custom}

\clearpage
\onecolumn
\appendix

\section{MUC-4 Template Slots}
\label{app:muc_slots}
Below is the complete list of \muc slots, which are the same for all template types, along with their definitions as provided in the conference appendices \citep{nn-1992-appendix}.\footnote{The original MUC-3 and MUC-4 data can be found at the following URL: \url{https://www-nlpir.nist.gov/related_projects/muc/muc_data/muc_data_index.html}. The licit set of values for each set-fill slot can also be found in \citep{nn-1992-appendix}. While the slots are the same across template types, the licit values of some set-fill slots are type-dependent.} The names of the string-fill slots are \textbf{bolded} and their (more commonly used) alternative names are given in parentheses. The significant majority of others are set-fill, though some slots require a numerical answer (e.g. ``PHYS TGT: NUMBER'') and these are known as \emph{text conversion} slots, as they require \emph{converting} possibly implicit counts of entities in the text into explicit numerical values. We group these with set-fill slots in the main text as they have likewise traditionally been excluded from evaluation since the original conference. ``MESSAGE: ID'' and ``MESSAGE: TEMPLATE'' were never part of the evaluation, even in the original conference. Some of the slot names use one or more of the following abbreviations: PERP = perpetrator; PHYS = physical; TGT = target; HUM = human.
\begin{enumerate}
    \item MESSAGE: ID --- The first line of the message, e.g., DEV-MUC3-0001 (NOSC). This slot serves as an index and is not scored in its own right.
    \item MESSAGE: TEMPLATE --- A number that distinguishes the templates for a given message. In the answer key, the word OPTIONAL in parentheses after the template number indicates that there is significant doubt whether the incident belongs in the database.
    \item INCIDENT: DATE --- The date of incident (according to local time, not Greenwich Mean Time).
    \item INCIDENT: LOCATION --- The place where the incident occurred.
    \item INCIDENT: TYPE --- A terrorist act reported on in the message.
    \item INCIDENT: STAGE OF EXECUTION --- An indicator of whether the terrorist act was accomplished, attempted, or merely threatened.
    \item \textbf{INCIDENT: INSTRUMENT ID } (\texttt{Weapon}) --- A device used by the perpetrator(s) in carrying out the terrorist act.
    \item INCIDENT: INSTRUMENT TYPE --- The category that the instrument fits into.
    \item PERP: INCIDENT CATEGORY --- The subcategory of terrorism that the incident fits into, as determined by the nature of the perpetrators.
    \item \textbf{PERP: INDIVIDUAL ID } (\texttt{PerpInd}) --- A person responsible for the incident.
    \item \textbf{PERP: ORGANIZATION ID } (\texttt{PerpOrg}) --- An organization responsible for the incident.
    \item PERP: ORGANIZATION CONFIDENCE --- The way a perpetrator organization is viewed in the message.
    \item \textbf{PHYS TGT: ID } (\texttt{Target}) --- A thing (inanimate object) that was attacked.
    \item PHYS TGT: TYPE --- The category that the physical target fits into.
    \item PHYS TGT: NUMBER --- The number of physical targets with a particular ID and TYPE.
    \item PHYS TGT: FOREIGN NATION --- The nationality of a physical target, if the nationality is identified in the article and if it's different from country where incident occurred.
    \item PHYS TGT: EFFECT OF INCIDENT --- The impact of the incident on a physical target.
    \item PHYS TGT: TOTAL NUMBER --- The total number of physical targets.
    \item \textbf{HUM TGT: NAME } (\texttt{Victim}) --- The name of a person who was the obvious or apparent target of the attack or who became a victim of the attack.
    \item \textbf{HUM TGT: DESCRIPTION} --- The title or role of a named human target or a general description of an unnamed human target.
    \item HUM TGT: TYPE --- The category that the human target fits into.
    \item HUM TGT: NUMBER -- The number of human targets with a particular NAME, DESCRIPTION, and TYPE.
    \item HUM TGT: FOREIGN NATION -- The nationality of a human target, if the nationality is identified in the article and if it's different from country where incident occurred.
    \item HUM TGT: EFFECT OF INCIDENT -- The impact of the incident on a human target(s).
    \item HUM TGT: TOTAL NUMBER -- The total number of human targets.
\end{enumerate}

\section{Data Collection}
\label{app:data_collection}
This appendix presents additional details about our data collection procedure, including the instructions that were provided to annotators (\S\ref{app:subsec:task_instructions}), screenshots of the annotation interface (\S\ref{app:subsec:task_interface}), and some measures and discussion of data quality (\S\ref{app:subsec:agreement}).

All annotators were told about the broad goals of the project prior to starting the task and were told that their annotations would be used for this project. All linguists who provided annotations are employees of the HLTCOE who receive a regular salary for annotation work, though we (the authors) were not informed of the exact salary of each annotator. Some of the native speaker annotators were authors of the paper and were not paid, as mentioned in \S\ref{sec:data_collection}; others were undergraduate students at Johns Hopkins, recruited through an internal job posting. The \$0.29 per-task pay rate given in the main text was computed by dividing the total pay for student annotators for each language (\$720) by the total number of tasks for each language (2,450). All annotation has been approved by Johns Hopkins.

\subsection{Task Instructions}\label{app:subsec:task_instructions}
Below are the task instructions that were presented to the annotators.

\subsubsection*{Overview}
In each task, a pair of sentences, one in English (``source'') and one in another (``target'') language will be shown to the user. The English sentence will be shown on the top half of the screen and an automatic translation of the English sentence into the target language will be shown on the bottom half. Both sentences will be segmented into words (``tokenized''). The task is to verify and correct alignments between highlighted spans of English text (each consisting of one or more words) and their translations in the target language. In each English sentence, there will typically be more than one span to align. The user needs to annotate the English spans word by word. By clicking on each English word, a \emph{suggested} span in the target language, based on an automatic (``default'') alignment between words in the English and target language sentences, is highlighted as the default answer on the target side (bottom of the screen). In some cases, you may also have the option to correct the target language translation as well.

\subsubsection*{Instructions}

\paragraph{The default alignment}
\begin{itemize}
    \item If you think the default alignment is correct (and the translation, if correcting the translation), simply press ``submit.''
    \item If you want to modify the default alignment, select the corresponding source span, modify the target span, and press ``submit.''
\end{itemize}

\paragraph{Aligning spans}
\begin{itemize}
    \item Only the source spans we are interested in are highlighted. All other words in the source sentence are greyed out.
    \item While ideally aligned spans in the target language will consist of contiguous sequences of words, it’s OK to select non-contiguous target words if appropriate.
    \item It may sometimes be the case either that (1) a word in the English does not have any clear analogue in the target language, or (2) a word in the target language does not have any clear analogue in English. In these cases, you can do one of two things.
    \begin{itemize}
        \item One possibility is to align the word without a clear analogue to a closely related word. For instance, ``happiness'' in English is translated in French as ``le bonheur,'' where ``le'' is a definite article, which is not used in the English. Here, we would align ``le'' to “happiness,” since it’s part of a multi-word expression that denotes the same thing as ``happiness'' does. In general, this solution should be preferred.
        \item Another possibility is to simply remove the word from the alignment. In general, this should be done only if the word is \emph{not} part of a multi-word expression (unlike ``le'' in ``le bonheur'' above) or seems like a translation error (that you cannot correct; see \textbf{Retokenizing the target sentence}).
    \end{itemize}
    \item As we are not experts in most of the languages we are annotating here, you will likely encounter other difficult alignment decisions we have not foreseen. When you first encounter such instances, try to formulate general rules that seem sensible to you and apply them consistently throughout the rest of your annotation.
\end{itemize}

\paragraph{Retokenizing the target sentence}
\begin{itemize}
    \item If you see the ``RE-TOKENIZE'' button on the target side, you are allowed to edit the target side text to correct the potential mistakes in automatic translation or word segmentation. When correcting translations, you should correct ALL text in the sentence that needs it---not just the tokens highlighted by the default alignments. You are allowed to edit or remove existing tokens, add new words, or split or merge the existing words to correct word segmentation. When retokenizing, each word or punctuation mark should go on its own line.
    \item If you make changes using ``RE-TOKENIZE,'' the suggested target spans will be automatically adjusted. In general, this adjustment should be correct: any words on the target side that you did not change should remain aligned to the correct word on the source side, even if you insert or delete other words. Of course, if you delete an aligned word on the target side, alignments to that word will be removed. Importantly, the same will be the case if you edit an aligned word, so you will have to realign any edited words. If you do make changes using ``RE-TOKENIZE,'' you should always double-check that the alignments are correct before submitting.
\end{itemize}

\paragraph{Mistakes}
\begin{itemize}
    \item Finally, if you make a mistake during annotation or encounter a technical problem in the interface, please try to note down the ID of the task you are working on at the time and inform us of the mistake or problem. The Task ID can be found in the top right corner of the screen (``Task ID: $\langle$\#$\rangle$''). \textbf{Please get in the habit of noting the task ID as soon as you accept it!}
    \begin{itemize}
        \item \textbf{NOTE}: We have noticed that some workers accidentally click the submit button after re-tokenizing, when they mean to click the \textbf{save} button (to save their new tokenization). Please try to avoid doing this, but tell us if you do.
    \end{itemize}
\end{itemize}

\subsection{Task Interface}\label{app:subsec:task_interface}
\begin{figure}
    \centering
    \includegraphics[scale=0.32]{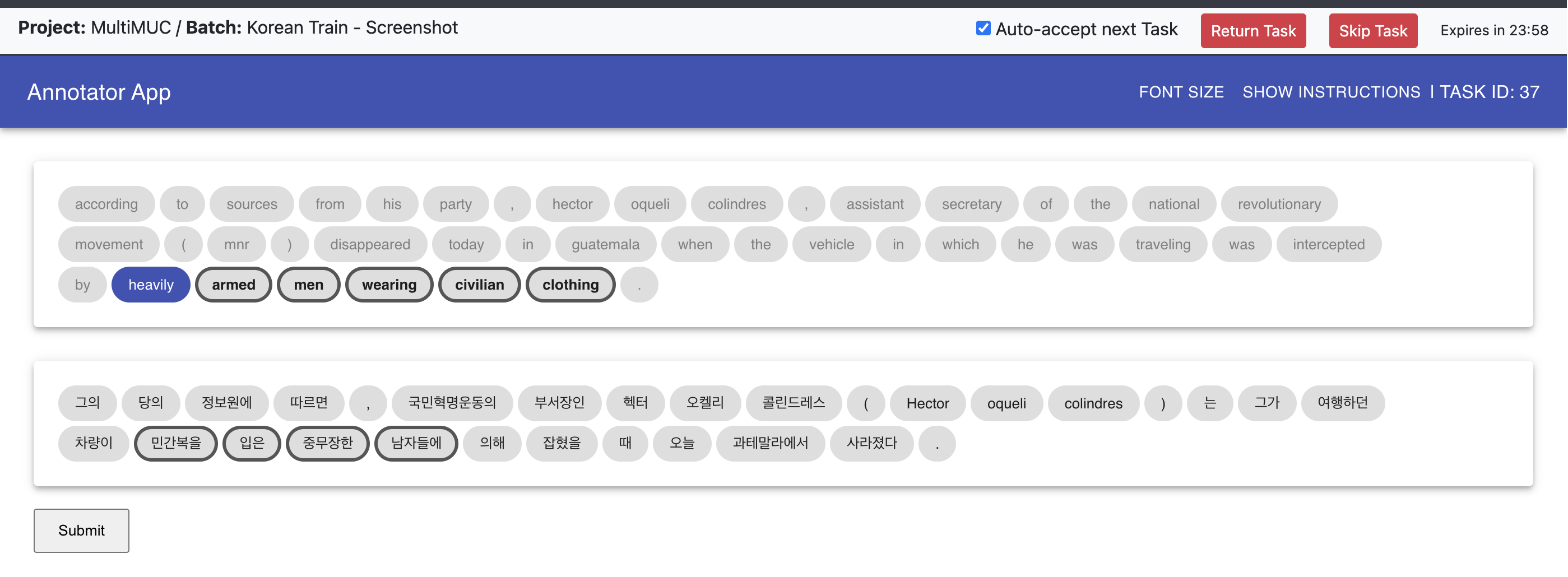}
    \includegraphics[scale=0.32]{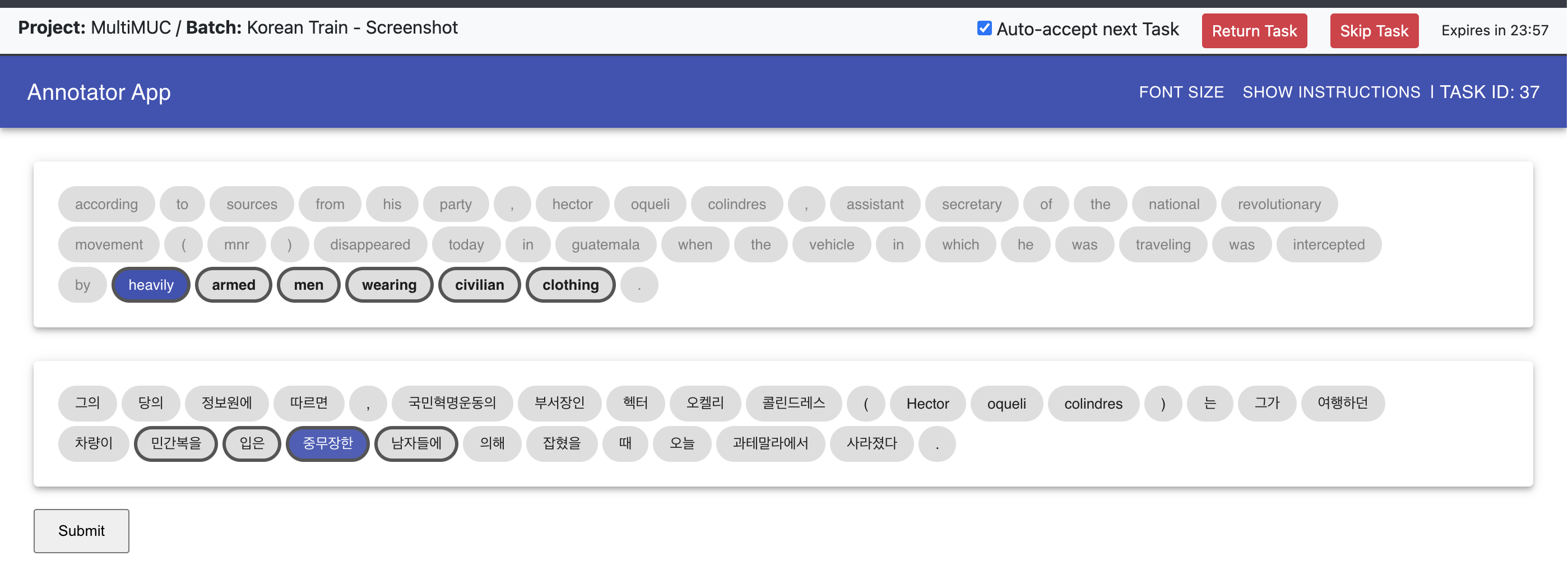}
    \caption{A Korean training split task before (top) and after (bottom) manual alignment correction.}
    \label{fig:kor-train}
\end{figure}

\begin{figure}
    \centering
    \includegraphics[scale=0.315]{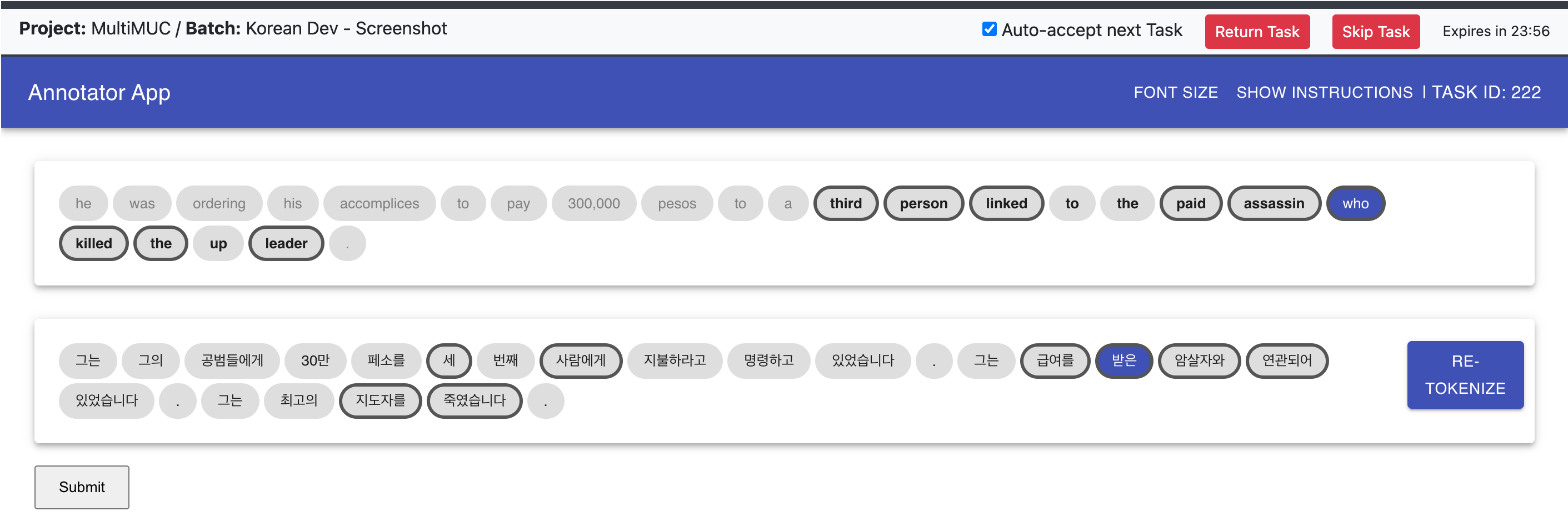}\\
    \includegraphics[scale=0.315]{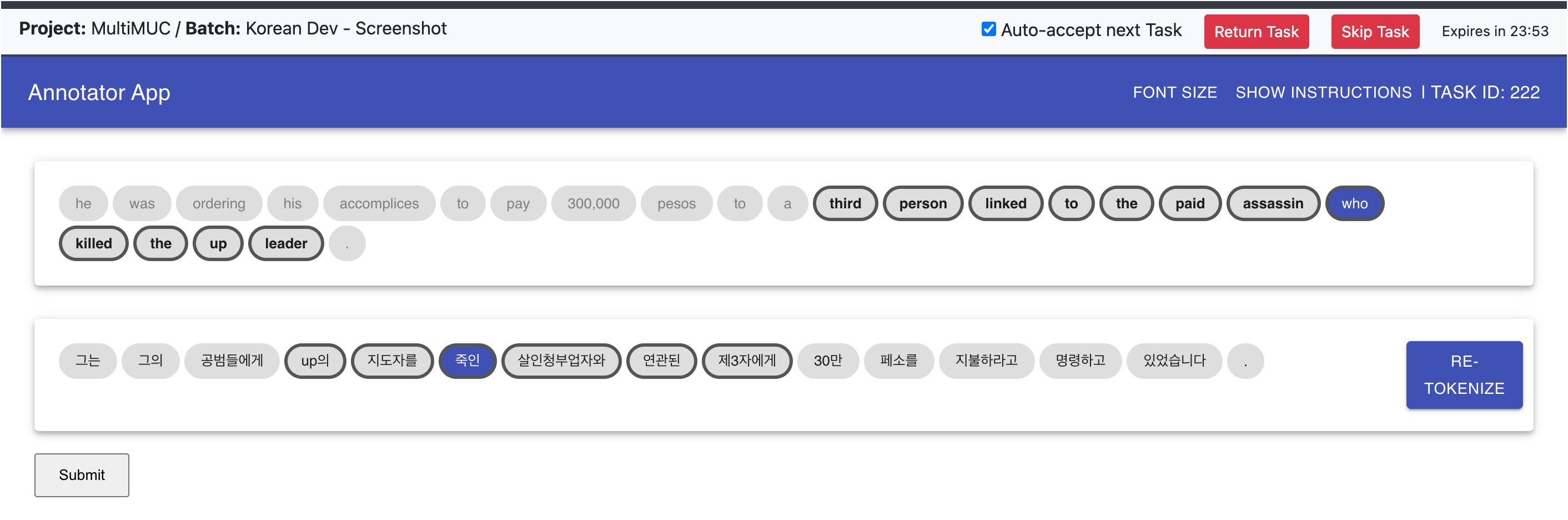}
    \caption{A Korean dev split task before (top) and after (bottom) manual alignment \emph{and} translation correction.}
    \label{fig:kor-dev}
\end{figure}

Recall from \S\ref{sec:data_collection} that alignment corrections were collected for all three splits (train, dev, and test) and that translation corrections were collected only for the dev and test splits. The same interface was used for both types of annotation. \autoref{fig:kor-train} and \autoref{fig:kor-dev} show examples of the interface for Korean annotation. \autoref{fig:kor-train} shows the interface as it appears when doing alignment correction only (i.e.\ training set annotation), both before any alignment correction (top) and after (bottom). \autoref{fig:kor-dev} shows the interface as it appears when \emph{also} doing translation correction (i.e.\ dev and test set annotation)---once again both before correction (top) and after (bottom). The only difference in the interface between the two figures is the presence of the ``RE-TOKENIZE'' button in \autoref{fig:kor-dev}, which, when clicked, allows annotators to change (insert/edit/delete) target language tokens. In both cases, when a new task is loaded, the annotator sees a ``default alignment,'' which is simply the automatic token alignment that is obtained using Awesome-align \citep{dou-neubig-2021-word} and that is in the \tgtauto experiments. This is the alignment they must correct (if necessary).


\subsection{Data and Annotation Quality}\label{app:subsec:agreement} As discussed in \S\ref{sec:data_collection}, our annotators were all either native speakers of the language they annotated or else were linguists with significant formal training in that language. Given this, and given that effective alignment and translation correction require only linguistic competence, the quality of the annotations can be presumed to be very high.

Even so, we provide some limited quantitative measures of annotation quality. We first report inter-annotator agreement on alignment correction for Farsi and Chinese for a randomly selected 50 tasks from the training set. We report Cohen's $\kappa$ at the token level: two alignments for a particular English token count as equivalent iff they align exactly the same target language token(s) to that English token. Two annotators completed these tasks for each language. For Farsi, we obtained a $\kappa$ of 0.98. For Chinese, we obtained a $\kappa$ of 0.87. Both indicate ``almost perfect'' agreement.\footnote{\url{https://en.wikipedia.org/wiki/Cohen\%27s_kappa\#Interpreting_magnitude}}

We additionally report sacreBLEU scores \cite{post-2018-call} between the uncorrected and corrected dev and test data for all languages to give a more quantitative sense of how similar the translation corrections are to the original, machine-translated text. The BLEU scores on the combined dev and test sets for Arabic, Farsi, Korean, Russian, and Chinese are (respectively) 73.1, 83.6, 76.1, 89.3, and 65.2. BLEU scores higher than 60 are often considered ``better than human''\footnote{\url{https://cloud.google.com/translate/automl/docs/evaluate\#interpretation}} and imply that the uncorrected and corrected translations can be considered as translations of the same source.

Finally, as we note in the limitations section, due to the lack of translation correction for the training set, translation errors resulted in alignment/projection failures for a small fraction of entity mentions. This included 4.6\% of mentions (and 3.2\% of entities) for Arabic, 3.0\% of mentions (2.4\% of entities) for Farsi, 4.4\% of mentions (3.1\% of entities) for Korean, 6.9\% of mentions (4.1\% of entities) for Russian, and 17.7\% of mentions (15.6\% of entities) for Chinese. We are in the process of correcting these cases and will be releasing a new version of the data with the corrections at \url{https://github.com/wgantt/multimuc}.

\section{Additional Results}
\label{app:additional_results}
As noted in \S\ref{sec:experiments}, we also considered a fourth setting for our supervised template filling experiments, \allman, which is similar to \biman except that models are trained on the gold English data and the corrected training data for \emph{all} \mm languages, using macro-average dev performance across languages for early stopping. \autoref{tab:additional_results} shows the results, with \biman numbers repeated from \autoref{tab:main_results}.

\gtt shows gains in CEAF-REE scores under the \allman setting for three languages (Arabic, Korean, Russian) and minor gains in CEAF-RME scores for Russian. In all other cases, however, \gtt's performance is comparable to, or somewhat lower than, what we observe in the \biman setting. Given these results, we do not think the greater compute requirements of the \allman setting are warranted.

The story is less ambiguous for \iterx, where we observe substantial performance degradations under the \allman setting relative to \biman. A significant part of the benefit that \biman confers on \iterx's performance  (relative to \tgtman and \tgtauto) is likely a consequence of \biman exposing the model to more English fillers, which occasionally appear untransliterated in the target language, as the additional English data in \biman may help the model learn to recover such fillers more accurately. However, it's very unclear what further benefits non-English, non-target language data could provide---especially given the diversity of language families represented here---and for \iterx, it seems only to confuse the model.

\begin{table*}[t]
    \centering
    \adjustbox{max width=\textwidth}{
    \begin{tabular}{ll|cccccc|cccccc}
    \toprule
    & & \multicolumn{6}{c}{\underline{CEAF-REE}} & \multicolumn{6}{c}{\underline{CEAF-RME}} \\
    & & En & Ar & Fa & Ko & Ru & Zh & En & Ar & Fa & Ko & Ru & Zh \\
        \midrule
    \multirow{2}{*}{\gtt} & \textsc{\biman} & & 36.76 & \textbf{37.91} & 36.52 & 36.97 & \textbf{41.48} & & \textbf{21.98} & \textbf{22.44} & \textbf{20.71} & 21.26 & \textbf{23.26} \\
    & \textsc{\allman} & & \textbf{37.77} & \textbf{37.91} & \textbf{37.31} & \textbf{38.63} & 37.11 & & 21.27 & 20.50 & 19.81 & \textbf{21.83} & 20.83 \\
        \midrule
    \multirow{2}{*}{\iterx} & \textsc{\biman} & & \textbf{34.73} & \textbf{33.15} & \textbf{37.02} & \textbf{36.95} & \textbf{36.02} & & \textbf{21.46} & \textbf{20.66} & \textbf{23.91} & \textbf{23.77} & \textbf{21.93} \\
    & \textsc{\allman} & & 20.98 & 28.92 & 21.53 & 27.64 & 28.94 & & \phantom{0}6.16 & \phantom{0}6.38 & \phantom{0}6.39 & \phantom{0}7.37 & 11.49 \\
    \bottomrule
    \end{tabular}
    }
    \caption{\iterx and \gtt results under the \biman and \allman settings (\biman results are repeated from \autoref{tab:main_results}). While we observe modest improvements in \gtt's CEAF-REE scores for some languages, most results suggest that bilingual training should be preferred (and for \iterx, strongly preferred) over joint training on all languages.}
    \label{tab:additional_results}
\end{table*}

\section{Training and Hyperparameters}
\label{app:training}
Our choices of hyperparameters for both \gtt (\S\ref{app:gtt}) and \iterx (\S\ref{app:iterx}) follow those associated with the best results in prior work (modulo a change in encoders) and are detailed below. While there is likely room for performance improvements from adopting language-specific encoders and hyperparameters, we leave these experiments for future work. The results for the models in the main text are based on single training runs, each of which was conducted on a single 24GB NVIDIA RTX 6000 GPU using the stopping criteria specified below. \S\ref{app:chatgpt} gives details on API hyperparameters and prompts for ChatGPT.

\subsection{GTT}\label{app:gtt}
We use the \gtt code base, available here: \url{https://github.com/xinyadu/gtt}. We use the hyperparameter settings exactly as listed in Appendix B of \citet{du-etal-2021-template}, with the following changes:
\begin{itemize}
    \item We used the cased version of mBERT-base \citep{devlin-etal-2019-bert} as the encoder in lieu of the original uncased BERT-base encoder.
    \item We train for 30 epochs in all experiments, as we found the default for \muc (18) to be insufficient for convergence in most cases. We use the checkpoint associated with best token-level accuracy on the dev set (this is the default behavior of GTT).
\end{itemize}
Since the \muc data is uncased, we also experimented with \emph{uncased} mBERT, though we found it yielded consistently worse performance. \citet{devlin-etal-2019-bert} in fact expressly recommend using the cased model, on the grounds that it corrects various issues with the uncased version.\footnote{See here: \url{https://github.com/google-research/bert/blob/master/multilingual.md}.}

\subsection{IterX}\label{app:iterx}
We use the \iterx code base, available here: \url{https://github.com/wanmok/iterx}. We use the same hyperparameters for \iterx as are listed in the ``best'' column of Table 7 in \citet{chen-etal-2023-iterative}, with the following changes:
\begin{itemize}
    \item We trained on \emph{gold} spans (rather than those predicted by an upstream system), as we empirically found this yielded superior results for \mm.
    \item We used mT5-base as the encoder to accommodate all \mm languages, as discussed in \S\ref{sec:experiments}.
\end{itemize}
\citeauthor{chen-etal-2023-iterative} report only average training time for \muc in their work, but we use the default maximum epochs (150) and patience (30) provided for the \muc training configuration in their repository. We limit total training time to 24 hours.

To ensure fair comparison across settings for inference (including for validation), we fix the candidate spans for all settings to those predicted for the relevant language by the span extraction system of \citet{xia-etal-2021-lome} that we trained for that language in the \biman setting (see \S\ref{subsec:span_extraction}).

\subsection{ChatGPT}\label{app:chatgpt}
The few-shot experiments described in \S\ref{subsec:few_shot} were run using \texttt{gpt-3.5-turbo-16k-0613} with a maximum context length of 8,192, a maximum of 1,024 new tokens to be generated, a temperature of 0.5, and a top $p$ of 1.0, with no presence penalty, frequency penalty, or logit biases. A single completion was generated per prompt. We recognize the potential for non-trivial performance variation that may result from even relatively minor changes to a prompt. Given the length of our prompts, cost prohibited us from running multiple variations for the main experiments, so results should be interpreted with caution.

The system prompt for all experiments was as follows:
\begin{quote}
You are an expert in information extraction, where you are given a few exemplars to help you understand the task. You have to perform textual analysis on a new document thereafter. Your analysis should be based on the ontology (inferred) and the exemplars.
\end{quote}
The structure of the remainder of the prompt is shown below, with prompt-specific components (i.e.\ the exemplars) described in italicized purple \pcomment{comments}. Each ``[DOCUMENT TEXT]:'' together with the full text document that followed constituted a single \textbf{user} message (provided as input in the \texttt{messages} API parameter). Likewise, each ``[TEMPLATES]:'' together with the annotated templates that followed constituted a single \textbf{assistant} message. The final instructions (``Please follow...'') and target document made up the last user message. All templates in the exemplars are formatted in the same way as the one given in the initial instructions below.

\begin{quote}
You are given a few exemplars to learn how to perform the template extraction task. You have to learn to do the same extraction to a new document. There are only 5 roles to use: PerpInd, PerpOrg, Target, Victim, Weapon.
Valid incident types are: ATTACK, ARSON, ROBBERY, BOMBING, KIDNAPPING, FORCED\_WORK\_STOPPAGE, BOMBING\_OR\_ATTACK, ATTACK\_OR\_BOMBING.
A target structures looks like this:
Template(incident\_type=``bombing'', PerpInd=[Entity(mentions=[Mention(``guerilla column'')])], PerpOrg=[Entity(mentions=[Mention(``army of national liberation''), Mention(``eln'')])], Target=[Entity(mentions=[Mention(``4-wheel drive vehicle''), Mention(``vehicle'')])], Victim=[Entity(mentions=[Mention(``carlos julio torrado'')]), Entity(mentions=[Mention(``torrado's son, william''), Mention(``william'')]), Entity(mentions=[Mention(``gustavo jacome quintero'')]), Entity(mentions=[Mention(``jairo ortega'')])], Weapon=[Entity(mentions=[Mention(``four explosive charges''), Mention(``explosive charges'')])])

[EXEMPLARS]:

[DOCUMENT TEXT]:

\pcomment{full text of example document 1 (least relevant; always in target language)}

[TEMPLATES]:

\pcomment{gold templates for example document 1 (always in target language)}

[DOCUMENT TEXT]:

\pcomment{full text of example document 2 (second most relevant; always in target language)}

[TEMPLATES]:

\pcomment{gold templates for example document 2 (always in target language)}

[DOCUMENT TEXT]:

\pcomment{full text of example document 3 (most relevant; in target language except in \biman setting)}

[TEMPLATES]:

\pcomment{gold templates for example document 3 (in target language except in \biman setting)}

Please follow the previous exemplars to process the new document. You have to use the same domain specific language to describe your extraction results. Do not add additional explanations except for the DSL generated. Make sure that you stick to the exact DSL as shown in the exemplars.

[DOCUMENT TEXT]:

\pcomment{full text of target (test set) document (always in target language)}

\end{quote}

\end{document}